\documentclass[sigconf,natbib=true]{acmart}

\AtBeginDocument{%
  \providecommand\BibTeX{{%
    \normalfont B\kern-0.5em{\scshape i\kern-0.25em b}\kern-0.8em\TeX}}}

\setcopyright{acmlicensed}
\copyrightyear{2024}
\acmYear{2024}
\acmDOI{XXXXXXX.XXXXXXX}

\acmISBN{978-1-4503-XXXX-X/18/06}
\usepackage[ruled,vlined]{algorithm2e}
\usepackage{fancyhdr}
\usepackage{caption}
\usepackage{graphicx}
\usepackage{subfigure}
\usepackage{colortbl}
\usepackage{amsmath}
\usepackage{booktabs}
\usepackage{multirow}

\begin{document}

\title{UniOQA: A Unified Framework for Knowledge Graph Question Answering with Large Language Models}



\author{Zhuoyang Li}
\affiliation{
  \institution{Xidian University}
  \city{Xian}
  \country{China}
  \postcode{710126}
}
\email{21009290002@stu.xidian.edu.cn}

\author{Liran Deng}
\affiliation{
  \institution{Xidian University}
  \city{Xian}
  \country{China}
  \postcode{710126}
}
\email{22031212323@stu.xidian.edu.cn}

\author{Hui Liu}
\affiliation{
  \institution{Xidian University}
  \city{Xian}
  \country{China}
  \postcode{710126}
}
\email{liuhui@xidian.edu.cn}

\author{Qiaoqiao Liu}
\affiliation{
  \institution{Xidian University}
  \city{Xian}
  \country{China}
  \postcode{710126}
}
\email{qiaoqiaoliu@stu.xidian.edu.cn}

\author{Junzhao Du}
\affiliation{
  \institution{Xidian University}
  \city{Xian}
  \country{China}
  \postcode{710126}
}
\email{dujz@xidian.edu.cn}
\authornote{Corresponding authors}

\renewcommand{\shortauthors}{Zhuoyang Li, Liran Deng, Qiaoqiao Liu et al. }


\begin{abstract}
  OwnThink stands as the most extensive Chinese open-domain knowledge graph introduced in recent times. Despite prior attempts in question answering over OwnThink (OQA), 
  existing studies have faced limitations in model representation capabilities, posing challenges in further enhancing overall accuracy in question answering. 
  In this paper, we introduce UniOQA, a unified framework that integrates two complementary parallel workflows. Unlike conventional approaches, UniOQA harnesses large language models (LLMs) for precise question answering and incorporates a direct-answer-prediction process as a cost-effective complement. 
  Initially, to bolster representation capacity, we fine-tune an LLM to translate questions into the Cypher query language (CQL), tackling issues associated with restricted semantic understanding and hallucinations. Subsequently, we introduce the Entity and Relation Replacement algorithm to ensure the executability of the generated CQL. 
  Concurrently, to augment overall accuracy in question answering, 
  we further adapt the Retrieval-Augmented Generation (RAG) process to the knowledge graph. Ultimately, we optimize answer accuracy through a dynamic decision algorithm. Experimental findings illustrate 
  that UniOQA notably advances SpCQL Logical Accuracy to 21.2\% and Execution Accuracy to 54.9\%, achieving the new state-of-the-art
  results on this benchmark. Through ablation experiments, 
  we delve into the superior representation capacity of UniOQA and quantify its performance breakthrough.
\end{abstract}

\ccsdesc[500]{Information systems~Question answering}
\ccsdesc[500]{Computing methodologies~Natural language processing}
\ccsdesc[500]{Information systems~Novelty in information retrieval}

\keywords{
Knowledge Graph Question Answering, large language model, Retrieval-Augmented Generation.
}



\maketitle

\section{Introduction}
\label{sec:introdction}
Knowledge Graph Question Answering (KGQA) aims to answer the natural language questions over knowledge graphs.
OwnThink\textsuperscript{}\footnote{Source from : https://github.com/ownthink/KnowledgeGraphData}, the largest Chinese open-domain knowledge graph with 140 million information triples, has been widely applied in sentiment dialogue generation\cite{qgdhsc} and machine reading comprehension\cite{mrc}. 
Similar to the knowledge graphs commonly used in KGQA, OwnThink has rich and high-quality knowledge. 
Moreover, OwnThink has fewer constraints on resource conditions and is more straightforward to operate since we can omit the tedious entity disambiguation\textsuperscript{} \footnote{ For the most commonly used knowledge graphs like Freebase\cite{freebase}, we require approximately 100GB of ultra-high memory, while OwnThink, being much smaller in scale, is suitable for general hardware conditions. Moreover, OwnThink directly stores the entities. After converting questions into logical forms, we can execute them directly, skipping the step of mapping entities in logical forms to corresponding IDs.}. 
Therefore, leveraging OwnThink to implement question answering is feasible and promising.

In recent years, some studies have explored question answering over OwnThink (Hereinafter referred to as OQA) and introduced the first dataset for OQA, named SpCQL\cite{SpCQL}. 
Nevertheless, these works have insufficient representation abilities, and it is difficult to further improve the overall accuracy of question answering. 

The insufficient representation capacity of a model refers to its difficulty in correctly extracting semantic information, entities, and relations from natural language questions.  
Previous execution-based methods have already exposed this pain point.
Some methods, whether based on fixed rules\cite{bert_cql} or trained neural networks\cite{SpCQL}, struggle to accurately understand the semantics of complex problems like multi-hop questions, suffering from non-execution issues due to syntactic and semantic errors in the generated logical forms Cypher (CQL).  
 Regarding these limitations, some methods\cite{in-context} leverage LLMs\textsuperscript{}\footnote{Large language models (LLMs) refer to pretrained models with a large number of parameters and a large amount of training data\cite{Talkingaboutllm}. 
They possess strong generalization and representation capabilities, leading to breakthroughs in many traditional tasks\cite{zhaoxin}. } with in-context learning (ICL)\cite{icl} and chain of thought (COT)\cite{cot} to achieve the accurate conversion from questions to CQL. 
Nevertheless, LLMs suffer from hallucinations\cite{Hallucination}, namely they may generate errors or unexpected information, limiting the effectiveness of conversion. 
Fine-tuning (FT) LLMs typically demonstrates powerful semantic understanding and can alleviate hallucination issues\cite{dechallu}.
Furthermore, due to the presence of ambiguous expressions in the questions, the generated CQL may contain entities or relations that are not present in the knowledge graph. 
Therefore, it is necessary to further replace the entities and relations to align them with the knowledge graph. 
 To our knowledge, most previous methods have overlooked this step, inevitably limiting the accuracy of question answering.

 Even if the execution-based methods can accurately understand the questions and achieve alignment of entities and relations, we still face another pain point: 
 previous execution-based methods follow the Text-to-CQL\cite{SpCQL} pattern, which may encounter errors in the conversion from questions to CQL. And the singularity of the pattern makes direct optimization difficult\cite{medical, mat}. 
    Therefore, we take a different approach to overcome this bottleneck. A cost-effective strategy is to combine direct-answer-prediction methods\cite{decaf}, where the models can directly search the knowledge graph for relevant answers based on the questions. 
  Although proven to be less accurate\cite{web2018}, the retrieval-based approaches are orthogonal to the execution-based methods. They offer an alternative form of answers as a supplement, which can enhances overall performance.

To address the two aforementioned pain points and further enhance the accuracy of OQA, we present a unified framework, UniOQA, which combines two complementary parallel workflows: \textit{\textsf{Translator}} and \textit{\textsf{Searcher}}. 
Specifically, we employ the former workflow as the primary engine by efficiently translating natural language questions into executable CQL. 
Simultaneously, we utilize the latter workflow as a supplementary compensator by directly retrieving information relevant to the questions to obtain another type of answers. 
Ultimately, we derive the final answers via the dynamic decision algorithm. 

The contributions of this work are threefold: 
  (1)  We present UniOQA, a unified framework for question answering over OwnThink. 
  Unlike previous mainstream approaches, we fine-tune a large language model (LLM)\textsuperscript{}\footnote{https://huggingface.co/EasonLiZZ/NL2Cypher-lora} to generate query language CQL, ensuring precise semantic understanding of the questions and alleviating the hallucinations of LLMs. 
   It is worth emphasizing that we innovatively align the knowledge graph by substituting entities and relations in the generated CQL, further enhancing the executability of CQL; 
   (2) Previous execution-based methods have limitations in further improving the accuracy of question answering. 
   We take a different approach by introducing a novel direct-answer-prediction process called GRAG. 
   This process directly searches for relevant information from the knowledge graph and allows the generative model to infer answers.
   (3) In our experiments on SpCQL, for comprehensive and rigorous evaluation, we additionally introduce three common metrics: precision, recall, and F1. The experimental results demonstrate that UniOQA improves SpCQL Logical Accuracy to 21.2\% and Execution accuracy to 54.9\%, 
  establishing that UniOQA achieves the new state-of-the-art performance on SpCQL. 
  Additionally, we achieve promising and competitive results on the introduced metrics.

\section{RELATED WORK}

\begin{figure*}
  \centering
  \includegraphics[width=\textwidth]{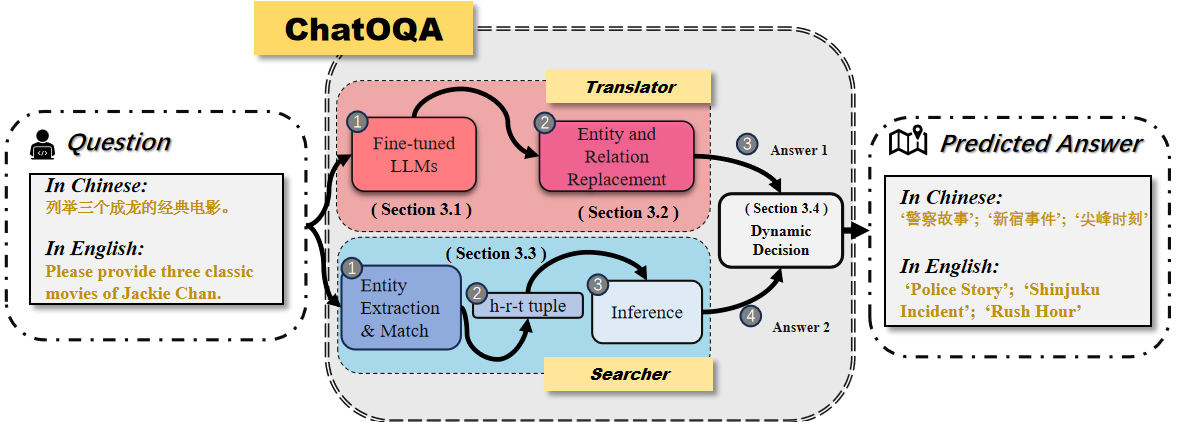} 
  \caption{The overview of our framework. We construct it with two parallel workflows: (1) \textit{\textsf{Translator}} (Section \ref{sec:ftllm}\&\ref{sec:algr}), which consists of fine-tuning LLM for  CQL generation and modifying the entities and relations in CQL. 
  (2) \textit{\textsf{Searcher}} (Section \ref{sec:grag}), which employs a direct search approach within the knowledge graph to retrieve answers relevant to the posed questions. 
  Finally, the answers from the two workflows undergo optimization via the dynamic decision algorithm (Section \ref{sec:comb}), yielding the ultimate results.   The example in this figure have been thoroughly derived in Section \ref{sec:met}.}

  \label{fig:flowchart}
\end{figure*}

\subsection{Text-to-CQL}
\label{sec:rlt2c}

OwnThink is advantageous due to its high-quality data, ease of operation and low resource constraints. 
A key approach to implementing question answering over OwnThink is to convert natural language questions into query language CQL and then execute the CQL to obtain the answers (i.e. Text-to-CQL\cite{SpCQL}). 
Several methods have already achieved it.

Firstly, some works focus on transforming questions into a sequence of words to match templates\cite{word_level_match} or extracting entities from questions and then generate CQL following certain rules\cite{bert_cql}.
However, fixed templates or rules lack generalization capabilities and flexibility, which potentially leads to significant errors in complex scenarios.
Regarding these limitations, some approaches have sought alternative solutions. 
Some neural networks are trained for directly transforming natural language questions into CQL, such as works\cite{SpCQL} based on Seq2Seq\cite{seq2seq}.
Compared to rule-based methods, deep neural networks have stronger generalization capabilities. 
Simultaneously, given the maturity of Text-to-SQL\cite{ersql,lgesql}, some studies\cite{sql2cql} transform natural language questions to SQL, then map to CQL.
Nevertheless, these approaches are still susceptible to factors such as poor network training or failed SQL conversion.
Some recent works\cite{in-context} are based on large language models (LLMs) with auxiliary programs and customized instructions. Nevertheless, many of these LLM-based works do not explicitly optimize model parameters, which lacks stability and often struggle to surpass methods that explicitly optimize parameters, such as fine-tuning (FT).

Therefore, we select the fine-tuning method, training model parameters on SpCQL\cite{SpCQL} and successfully open-sourcing the fine-tuned model. 

In addition, the process of Text-to-CQL involves directly executing the generated logical forms to obtain answers, which is challenging to ensure the accuracy.
While in similar tasks, such as Text-to-SPARQL\cite{chatkbqa,decaf}, the generated logical forms require entity disambiguation to ensure executability\textsuperscript{} \footnote{Entity disambiguation is not necessary for Text-to-CQL, but it is indispensable for Text-to-SPARQL}.

Inspired by this step, we propose the Entity and Relation Replacement (ERR) algorithm to further improve the accuracy of Text-to-CQL.

\subsection{Retrieval-Augmented Generation}
Retrieval-Augmented Generation (RAG)\cite{RAG} enhances the performance of generative models by integrating retrieval information, which enables them to better understand and utilize information from external knowledge bases.

However, generative models may be affected by contextual length limitations and redundant information interference when using retrieved contextual information for inference. 
To overcome these challenges, Yang et al. train an information extractor to generate a compressed context of the input document as the output sequence\cite{prca}, while RECOMP employ a similar approach by using contrastive learning to train an information condenser to extract key information\cite{recomp}. 
However, the method of information compression can lead to issues of information loss and selection bias.
Unlike them, Zhuang et al. train a reordering model to prioritize the top-most relevant items\cite{Zhang}, providing more effective and accurate input for subsequent language model processing.  However, this reordering will bring additional computational costs.

In this paper, we retrieve relevant subgraphs of topic entities as context information, balancing the interference caused by context length and additional information.

\subsection{Joint Execuation Answers and Retrieval Answers}
\label{sec:decafjoint}
Our framework combines two powerful workflows, effectively enhancing the efficiency of question answering. Yu et al.\cite{decaf} propose a similar joint framework for tasks related to the Freebase\cite{freebase} knowledge base. 
However, they notably employ the BM25\cite{bm25} for knowledge base retrieval and utilize Fusion-in-Decoder\cite{fusion} to generate logical forms.
 Additionally, they ultimately adopt a binary logic combination rule (BNA), i.e., only if the execution results is empty, they use the retrieval results. 
 However, we believe that using the empty execution result as a condition weakens the role of the retrieval method. 
 Our combination rule (DDA) equally considers the execution and retrieval results by selecting the better ones as the answers to the questions (Section \ref{sec:comb}).

\section{METHODOLOGY}
As shown in Figure \ref{fig:flowchart}, our framework comprises two parallel workflows. The \textit{\textsf{Translator}} workflow involves fine-tuning a language model for CQL generation (Section \ref{sec:ftllm}) and modifying entities and relations in CQL ( Section \ref{sec:algr}). 
 The \textit{\textsf{Searcher}} workflow utilizes a direct search approach within the knowledge graph to retrieve answers (Section \ref{sec:grag}). Finally, answers from both workflows undergo optimization via the dynamic decision algorithm (Section \ref{sec:comb}).

\label{sec:met}
\subsection{CQL Generation by Fine-tuned LLMs}
\label{sec:ftllm}
Instruction-tuning is a method for fine-tuning on a collection of instances in natural language format\cite{instruction}, closely related to supervised fine-tuning\cite{zhaoxin}. 
We construct instructions to fine-tune LLMs in a supervised manner for CQL generation. 

We introduce a random noise vector, which is generated by sampling uniformly within the range of -1 to 1\cite{neft} in the embedding layer. 
After fine-tuning various LLMs, including Baichuan2-7B\cite{baichuan}, Atom-7B\textsuperscript{}\footnote{Atom is a variant of LLaMA2, which has been further enhanced on Chinese pre-training datasets. https://github.com/LlamaFamily/Llama-Chinese?tab=readme-ov-file}, Qwen\cite{Qwen}, ChatGLM3-6B\cite{chatglm}, Internlm-7B\cite{internlm} and ChatGPT(\texttt{gpt-3.5-turbo})\cite{gpt3}, 
experimental results (Table \ref{tab:main}) indicate that fine-tuned Baichuan2-7B exhibits superior performance. Consequently, we choose it to generate CQL.

To pose a trade-off between the effectiveness and hardware resource, we quantize the model weights to 4 bits\cite{qlora}, reducing the memory consumption required for deployment and inference. Simultaneously, UniOQA explores various Parameter Efficient Fine-Tuning (PEFT) methods\cite{peft} to efficiently fine-tune and reduce our resource consumption. Among them, we choose LoRA\cite{lora}, which freezes the pretrained model weights and injects trainable rank decomposition matrices into each layer of the Transformer architecture,
greatly reducing the number of trainable parameters by 1,0000 times and the GPU memory requirement by 3 times.

LLMs can perform preliminary transformations from natural language questions to CQL without correct entities and relations.
For the example in Figure \ref{fig:flowchart}, the fine-tuned LLM can generate CQL like: \texttt{"match(:ENTITY\{name:"Jackie Chan"\})-[:Relationship\{na-\\me:"classic movie"\}]->(m) return distinct m.name limit 3"}. However the correct entity in the knowledge graph is \texttt{"Jackie Chan [Hong Kong actor]"} rather than \texttt{"Jackie Chan"}.  It is difficult to generate the correct entity without any prompts. 
The same applies to relations. CQL generated solely by the model may not be entirely correct, hence the need for entity and relation replacement in the next section.

\subsection{Entity and Relation Replacement}
\label{sec:algr}
Based on the work in Section \ref{sec:ftllm}, LLMs can generate CQL with correct grammar and syntax. 
 However, most previous works stop here. They overlook the alignment with knowledge graphs, relying solely on extracting raw information from questions. 
To address this deficiency, we further propose the Entity and Relation Replacement (ERR) algorithm . 

We define the related variables as follows:
\begin{table}[h]
  \label{tab:var}
  \begin{tabular}{lcc}
    \toprule
    Variable & Description\\
    \midrule
    Q & Natural language questions  \\
    $CQL_p$ & The CQL with original entities and relations\\
    $CQL_f$  & The CQL with corrected entities and relations. \\
    \bottomrule
  \end{tabular}
\end{table}

\begin{algorithm}
  \label{tab:alg}
  \SetAlgoLined
  \caption{Entity and Relation Replacement}
  
  \KwIn{A list of natural language questions and original CQL generated from LLM \{$Q\text{:}C_p$\}, top-$k$ parameter, basemodel $Baichuan2-7B$, the entity set $S_E$ and the relation set $S_R$ of knowledge graph }
  \KwOut{The corrected list \{$Q\text{:}C_f$\}}
  
  \ForEach{$Q,CQL_p \in \{Q\text{:} C_p\}.items()$}{
  $E \leftarrow RegularlyExtract(CQL_p,"\text{:}Entity\{name\text{:}\}")$;\\
  $R \leftarrow RegularlyExtract(CQL_p,"\text{:}Relationship\{name\text{:}\}")$;\\  
  $E' \leftarrow \emptyset$;\\
  
  \ForEach{$e \in E$}{
    $E_b \leftarrow FindRelative(e,S_E)$;\\
    $e' \leftarrow LLMselect(E_b, Baichuan2\text{-}7B,Q)$;\\
    $E' .append(e')$;\\
  
    }
  
  $R' \leftarrow \emptyset$;\\
  $Score \leftarrow \emptyset$;\\
  $r \leftarrow R[0]$;\\
  $R_b \leftarrow FindRelative(r,S_R)$;\\
  
  \ForEach{$r_b \in R_b$}{
    $Score.append(SimiScore(r_b,r))$
  
    }
  
  $R' \leftarrow GetTopk(Score,k)$;\\
  $C \leftarrow \emptyset$;\\
  $CQL' \leftarrow Replace(E,E',CQL_p, Entity)$;\\
  
  \ForEach{$r' \in R'$}{
      $CQL_f \leftarrow Replace(r,r',CQL', Relationship)$;\\
      $C.append(CQL_f)$;\\
  
  }
  \If{$C$ is not empty}{
        $Best\text{-}CQL_f \leftarrow MaxAcc(C)$;\\
  }
  \{$Q\text{:}C_f\}.append(\{Q:Best\text{-}CQL_f\})$
}
\Return \{$Q\text{:}C_f$\}
\end{algorithm}

The ERR algorithm essentially replaces entities and relations in CQL with the most semantically similar entities and relations from the knowledge graph. 

As shown in Algorithm 1, the input is the original list \{$Q\text{:}C_p$\} and the output is the corrected list \{$Q\text{:}C_f$\}.
We iterate through each $Q$ and $CQL_p$ pair in sequence and utilize regular expressions for entity and relation extraction to form the entity set \textit{\textsf{$E$}} and relation set \textit{\textsf{$R$}}.

For each entity \textit{\textsf{$e$}} in \textit{\textsf{$E$}}, we retrieve all related entities \textit{\textsf{$S_E$}} from knowledge graph to form a candidate set: \textit{\textsf{$E_b \leftarrow \text{FindRelative}(e, S_E)$}}. Then, we 
leverage $Baichuan2\text{-}7B$ with a manually crafted instruction to select the final entity: \textit{\textsf{$e' \leftarrow \text{LLMselect}(E_b, $Baichuan2\text{-}7B$, Q)$}}. Finally, we obtain the corrected entity set $E'$.

Correcting the first relation $r$ in $R$, we obtain a candidate relation set: \textit{\textsf{$R_b \leftarrow \text{FindRelative}(r, S_R)$}}. 
For each element $r_b$ in $R_b$, we calculate the semantic similarity\textsuperscript{}\footnote{For the relations, we use a pre-trained tool, Synonyms, to calculate similarity. Source of Synonyms: https://github.com/chatopera/Synonyms}: \textit{\textsf{$Score.append(SimiScore(r_b,r))$}} and select the top k: \textit{\textsf{$R' \leftarrow GetTopk(Score,k)$}}. 
We obtain the corrected relation set $R'$. 

We reconstruct the CQL to obtain the formal CQL set $CQL_f$.
Finally, we select the best formal CQL based on execution accuracy: \textit{\textsf{$Best\text{-}CQL_f \leftarrow MaxAcc(C)$}}, and output the corrected pair list \{$Q\text{:}C_f$\}. 

Through the algorithm described above, we modify the incorrect entity from Section \ref{sec:ftllm}. By executing the modified CQL, we obtain the final answers of \textit{\textsf{Translator}}: "\texttt{Police Story}", "\texttt{Rush Hour}", "\texttt{Shinjuku Incident}". 
Although the answers in this example are correct, the ERR algorithm also has certain limitations, such as erroneously selecting the best entities or relations. 
Therefore, we introduce the work in the next section to obtain candidate answers.

\subsection{GRAG Process}
\label{sec:grag}
\begin{figure}[h]
  \centering
  \includegraphics[width=\linewidth]{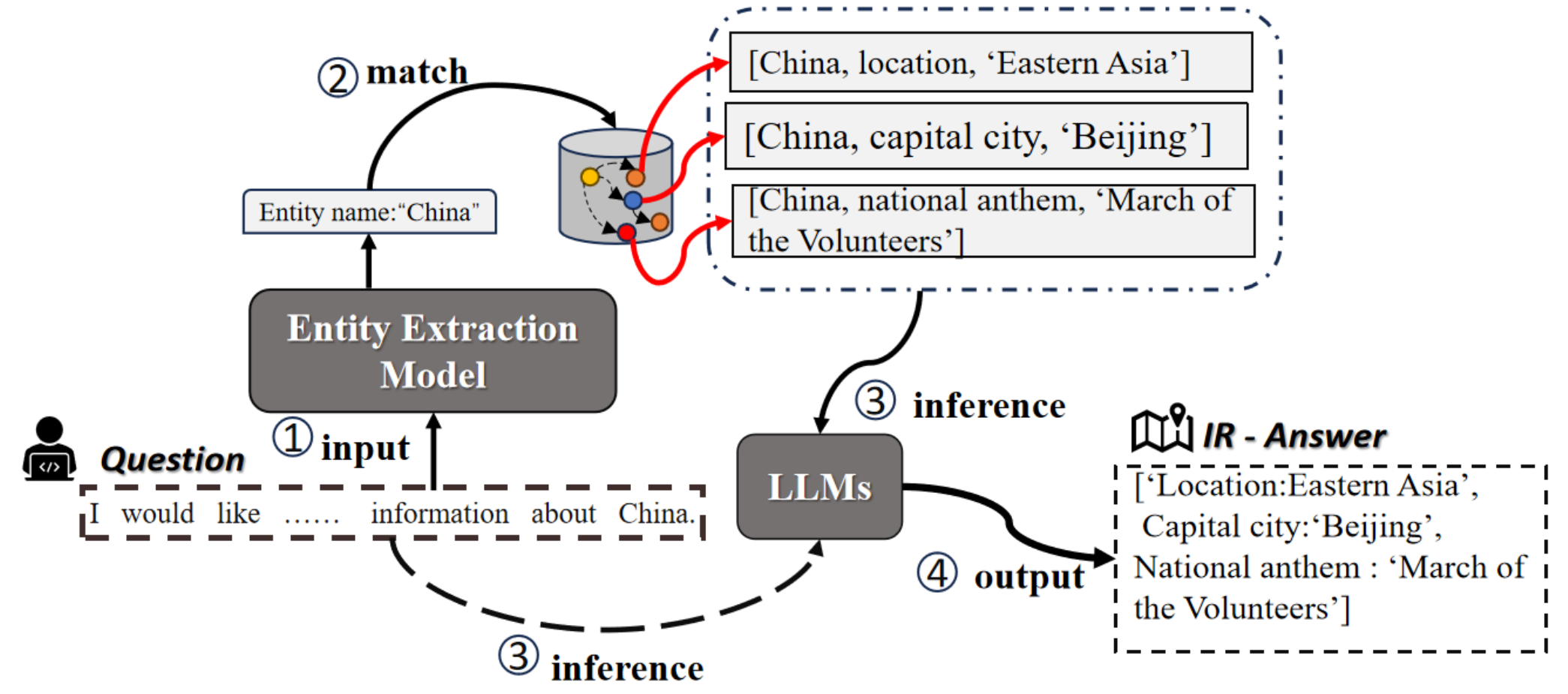}
  \caption{The overview of GRAG. Also the overview of the \textit{\textsf{Searcher}} workflow.}
  \Description{3 workflow in previous that generates wrong CQL.}
  \label{sec:fig1}
\end{figure}

As shown in Figure \ref{fig:flowchart}, we concurrently execute the \textit{\textsf{Searcher}} workflow. 
The completion of the \textit{\textsf{Translator}} workflow initially implement question answering, but we struggle to gain a more comprehensive understanding of natural language questions and to generate correct CQL.
In order to overcome this bottleneck, we introduce GRAG, which stands for applying the Retrieval-Augmented Generation framework to knowledge graphs, to directly retrieve the answers. (see Figure \ref{sec:fig1} for an illustration) .

We employ traditional information retrieval methods\cite{graphe} to retrieve subgraphs relevant to the topic entities as the context for LLMs. 
Firstly, we train an entity extraction model to extract entities from the question (e.g., for the question: \texttt{"I want to know some information about China",} the entity in the question is \texttt{"China"}). 
After obtaining the entities, to compensate for the potential inaccuracies in entity extraction, we use a pattern-matching query to retrieve triples from the Elasticsearch database \textsuperscript{}\footnote{Source from: https://github.com/elastic/elasticsearch} 
that are within one hop of the entity as the context information. 
The retrieved knowledge is presented in the form of triples [head, relation, tail], where each triple represents an implicit expression of a reasoning path. 
These aggregated triples are then transformed into natural language using templates (e.g., \texttt{"The relation of head is tail."}). 
Subsequently, the knowledge in natural language format is merged with the questions and input into a LLM\textsuperscript{}\footnote{We infer the answer by synthesizing all the triplet information via a large language model named InternLM-7B\cite{internlm}, which has demonstrated outstanding performance on natural language understanding tasks.}. 
Finally, the LLM is prompted to generate an answer based on the provided external knowledge. 
During the inference phase, we also explicitly prompt the model to generate the results in the form of labels.

Now, we complete the \textit{\textsf{Searcher}} workflow. Following the aforementioned process, we obtain another type of answers for the example shown in Figure \ref{fig:flowchart}: "\texttt{Police Story}", "\texttt{Rush Hour}", "\texttt{Shinjuku Incident}". 
In the next section, we will demonstrate how to combine the two types of answers we have obtained in Section \ref{sec:algr} and this section.

\subsection{Dynamic Decision Algorithm}
\label{sec:comb}
Our framework combines the outputs of the two aforementioned workflows. We prioritize the answers from \textit{\textsf{Translator}} while using the answers from \textit{\textsf{Searcher}} as a supplement. 
Through the formula we design, we dynamically combine two types of answers to obtain better results. The final answers for the example in Figure \ref{fig:flowchart} are "\texttt{Police Story}", "\texttt{Rush Hour}", "\texttt{Shinjuku Incident}".

 The specific formula is as follows:
\begin{equation}
  \text{Final Answer} = 
  \begin{cases}
      S(Q) & \text{if } F\text{1}(S(Q)) \geq \sigma , \\
      Better(S(Q),I(Q)) & \text{else}
  \end{cases}
  \end{equation}

where Q refers to the natural language question, $S(\cdot)$ indicates the answers of the questions from \textit{\textsf{Translator}}, $F1(\cdot)$ indicates the F1 score of the answers, $I(\cdot)$ indicates the answers from \textit{\textsf{Searcher}}. 
$Better(a, b)$ refers to selecting the answers with higher F1 score between $a$ and $b$. 
And $\sigma$ is a decision factor within the range of 0 to 1, determining the threshold of decision. 
Based on the experimental results(Figure \ref{fig:abla}(b)), we set it to 1, which means that we will select the superior answers as the final results.

\section{EXPRIMENTAL SETUP}

\subsection{Dataset}
We evaluate our approach on the SpCQL\cite{SpCQL}, which is constructed based on OwnThink. The dataset comprises over 10,000 question-CQL-answer triples, with a training-validation-test ratio of 7:1:2.
The questions in the dataset cover multiple topics, such as events, personalities, etc. Moreover, nearly 60\% of the CQL queries are relatively complex, involving conditional clauses like COUNT(*), WHERE, as well as multiple entities or relations.

To the best of our knowledge, SpCQL is the first and currently the only KGQA dataset built on OwnThink. It is crucial to note that commonly used KGQA datasets, such as FreebaseQA\cite{freebaseqa}, WebQSP\cite{webqsp} and ComplexWebQuestions\cite{cwq}, are built on knowledge graphs like Freebase\cite{freebase}. 
However,  our approach is specifically tailored to the OwnThink knowledge graph.

\subsection{Baseline Models}
It should be noted that the work on SpCQL is limited. Moreover, previous methods\cite{word_level_match, bert_cql,medical} are not available, or specific to other knowledge graphs\cite{sql2cql,mat}, which is hard to migrate to OwnThink. 
Therefore, it is difficult to directly utilize previous methods to evaluate our UniOQA and we attempt to select other representative methods as baselines.

We retain the existing baselines, which include Seq2Seq\cite{seq2seq}, Seq2Seq+Attention\cite{seq2seqattn}, Seq2Seq+Copy\cite{seq2seqcopy}.
Then, following previous empirical practice\cite{bird,nl2gql,nl2spl1,nl2spl2}, we present the performance of two types of baseline models in SpCQL. 
The first type of model is based on fine-tuning (FT), which outputs CQL by tuning parameters of language models to learn from the annotated train set. 
On the other hand, the second type of model is based on in-context learning (ICL) or chain-of-thought (COT), which generate results without additional training. 
Specifically, in ICL, input-output pairs are used to construct prompts, while COT incorporates intermediate reasoning steps into the prompts.
For FT models\textsuperscript{}\footnote{For our baselines, we primarily fine-tune the 7B-parameter LLMs(\texttt{gpt-3.5-turbo} may also have a 7B-parameter configuration\cite{gpt_only_7b}). For a fair comparison, we do not fine-tune GPT-4, which has a much larger parameter scale than 7B.},
 we select Baichuan2-7B\cite{baichuan}, Atom-7B\cite{llama}, Qwen-7B\cite{Qwen}, ChatGPT(\texttt{gpt-3.5-turbo}), ChatGLM3-6B\cite{chatglm} and InternLM-7B\cite{internlm}. 
For ICL-based models, we provide one-shot and few-shot results of ChatGPT (\texttt{gpt-3.5-turbo}) and GPT-4 (\texttt{gpt-4-turbo}). 
 In addition, we also compare the performance of UniOQA leveraging the previous combination algorithm (BNA) in Section \ref{sec:decafjoint}
  with UniOQA leveraging our dynamic decision algorithm (DDA) in Section \ref{sec:comb}.

\subsection{Metrics}
The benchmark includes the following two metrics:

{\bfseries (1) Logical Accuracy ($\textbf{ACC}_{\textbf{LX}}$)}
compares the CQL generated by the model with the logical form of {\bfseries gold CQL }\textsuperscript{}\footnote{The gold CQL refers to the correct CQL ,typically annotated manually by experts or obtained through other reliable methods}.
It may contain false positives caused by conditional order\cite{SpCQL}.

\begin{equation}
  \begin{aligned}
  ACC_{LX} &= \frac{\text{number of correctly predicted CQL}}{\text{total number of CQL}}
  \end{aligned}
  \end{equation}

{\bfseries (2) Execution Accuracy ($\textbf{ACC}_{\textbf{EX}}$)}
compares the execution results of CQL generated by the system with the execution results of {\bfseries gold CQL}.

\begin{equation}
  \begin{aligned}
  ACC_{EX} &= \frac{\text{number of CQL with correct execution result}}{\text{total number of CQL}}
  \end{aligned}
\end{equation}

Logical Accuracy can be underestimated because CQL with different logical forms may produce the same execution results. Differences in entities and relations can also contribute to this. Execution Accuracy, which requires a complete match with the correct answers, may not fully reflect the actual effectiveness of a model.

Therefore, for a more reasonable and comprehensive analysis, we additionally introduce three common metrics in previous works\cite{sp,kbqasur,neuralsym}for further evaluation, including:

{\bfseries (3) Precision (${\mathbf{P}}$)}  
reflects the the proportion of correct predictions in each predicted answer on average. It is calculated as:

\begin{equation}
  \begin{aligned}
    P = \frac{\left| \mathcal{A}_{g_i}\cap \mathcal{A}_{p_i} \right|}{\left| \mathcal{A}_{p_i}   \right|}
  \end{aligned}
\end{equation}

where $\mathcal{A}_{g_i}$ is the gold answer set for the i-th example,  $\mathcal{A}_{p_i}$ is the predicted answer set for the i-th example, ${P}$ is the precision of this example.

{\bfseries (4) Recall (${\mathbf{R}}$)}
reflects the proportion of correctly predicted answers out of the total answers in each standard answer on average. It is calculated as: 

\begin{equation}
  \begin{aligned}
    R  &= \frac{\left| \mathcal{A}_{g_i}\cap \mathcal{A}_{p_i} \right|}{\left| \mathcal{A}_{g_i}   \right|}
  \end{aligned}
\end{equation}

{\bfseries (5)  F1 (${\mathbf{F1}}$)} 
is the harmonic mean of precision and recall  for each predicted result on average.

\begin{equation}
  \begin{aligned}
  {F1}  &= \frac{2*P*R}{P+R}
  \end{aligned}
\end{equation}

\subsection{Experimental Setting}

For the experiments on ChatGPT and GPT-4, we utilize the API provided by OpenAI. As one of the most iconic models, we conduct fine-tuning with the same training steps as the aforementioned training set.
For the few-shot experiments, given the significant impact of the provided examples, we generate a total of $3$ runs and report the average results.

For the experiments on other LLMs, we train on SpCQL for 4000 steps, with a training batch size of 2, learning rate of 1e-4, warmup steps of 100, weight decay of 0.05, and neftune noise alpha of 5. 
We set AdamW\cite{admaw} as optimizer. We conduct our experiments on a single NVIDIA RTX 4090 (24GB) GPU hosted on the Autodl\textsuperscript{}\footnote{https://www.autodl.com/home}.

\section{EXPERIMENTAL RESULTS}
In this section, we investigate the effectiveness of UniOQA. 
Firstly, we compare our approach with the baselines on the SpCQL\cite{SpCQL} dataset,
 demonstrating that our approach achieves the new state-of-the-art results on SpCQL.  
 Next, we conduct ablation studies to further analyze the superiority of UniOQA
  in representation capacity and quantify its breakthrough in performance.

 \subsection{Main Result}

 \begin{table}[htbp]
   \caption{The experimental results on SpCQL(\%)}
   \label{tab:main}
   \begin{tabular}{lcccccc}
     \toprule
     & \multicolumn{5}{c}{\textbf{SpCQL}}  \\ \cline{2-6} 
     \multirow{-2}{*}{\textbf{Model}}& $\mathbf{ACC_{LX}}$ & $\mathbf{ACC_{EX}}$ & $\mathbf{P}$ & $\mathbf{R}$ & $\mathbf{F1}$ \\

     \midrule
     \multicolumn{6}{c}{\cellcolor{gray!30}\textbf{Existing Baselines}} \\
     Seq2Seq & $1.7$ & $1.9$ & \text{/} & \text{/} & \text{/} \\
     Seq2Seq+Attention & $1.8$ & $2.0$ & \text{/} & \text{/} & \text{/} \\
     Seq2Seq+Copy & $2.3$ & $2.6$ & \text{/} & \text{/} & \text{/} \\
     \multicolumn{6}{c}{\cellcolor{gray!30}\textbf{ICL/COT}} \\
     ChatGPT+one-shot &/& $3.3$ &$5.3$ &$5.3$&$4.9$\\ 
     ChatGPT+cot &/&$3.5$&$5.4$&$5.4$&$4.9$\\
     ChatGPT+few-shot & / & $11.5$ & $13.7$ & $14.1$ & $13.5$ \\
     GPT-4+few-shot & / & $12.6$ & $16.5$ & $16.4$ & $16.0$ \\
     \multicolumn{6}{c}{\cellcolor{gray!30}\textbf{FT}} \\
     InternLM-7B & $/ $ &$27.7$ & $35.3$ & $34.5$ & $34.2$\\
     ChatGLM3-6B & $/ $ &$32.0$ & $40.9$ & $42.3$ & $40.1$\\
     ChatGPT & $16.6 $ &$41.3$ & $50.6$ & $50.7$ & $49.6$\\
     Qwen-7B & $19.6 $ &$47.6$ & $57.8$ & $58.0$ & $56.8$\\
     Atom-7B & $20.4 $ &$48.9$ & $58.0$ & $58.6$ & $57.2$\\
     Baichuan2-7B & $20.9$ &$48.9$ & $58.8$ & $59.4$ & $58.0$\\
     \midrule
     UniOQA+BNA & $21.2$ &$53.8$ & $66.3$ & $69.0$ & $65.8$\\
     \textbf{UniOQA+DDA}(ours) & $\textbf{21.2} $&$\textbf{54.9}$ & $\textbf{67.1} $& $\textbf{69.2}$&$\textbf{66.3}$\\
     \bottomrule
   \end{tabular}
 \end{table}
 
 Table \ref{tab:main} presents the experimental results for our proposed UniOQA and other baseline models. 
 We observe a significant improvement of UniOQA compared to existing works on SpCQL. $ACC_{LX}$ and $ACC_{EX}$ have increased to 21.2\% and 54.9\%, achieving the new state-of-the-art results on the SpCQL dataset. 
 Additionally, UniOQA outperforms various methods based on large language models (LLMs) in all metrics. 
  
 Recall in Section \ref{sec:introdction} and Section \ref{sec:rlt2c}, we discuss the effective approaches to overcoming the hallucination problem in LLMs. 
 We argue that methods such as in-context learning (ICL) and chain of thought (COT), which do not explicitly optimize the models, can lead to inaccuracies in generating CQL, while fine-tuning (FT) methods train the model's parameters, 
 resulting in more accurate CQL generation. 
 As shown in Table \ref{tab:main}, all FT-based methods are significantly better than ICL-based and COT-based methods, confirming our hypothesis. 
 
 Furthermore, different LLMs exhibit significant performance differences on SpCQL, which we attribute to differences in model architecture and pretraining data. 
 The well-performing Baichuan2-7B and Atom-7B both adopt a decoder-only architecture, demonstrating stronger zero-shot capabilities\cite{bloom}, which are crucial indicators of model performance. 
 Additionally, we find in the Baichuan2-7B report\cite{baichuan} that there is a significant proportion of pretraining data related to code, which may also positively influence our task.

 \begin{figure}[h]
  \centering
  \includegraphics[width=\linewidth]{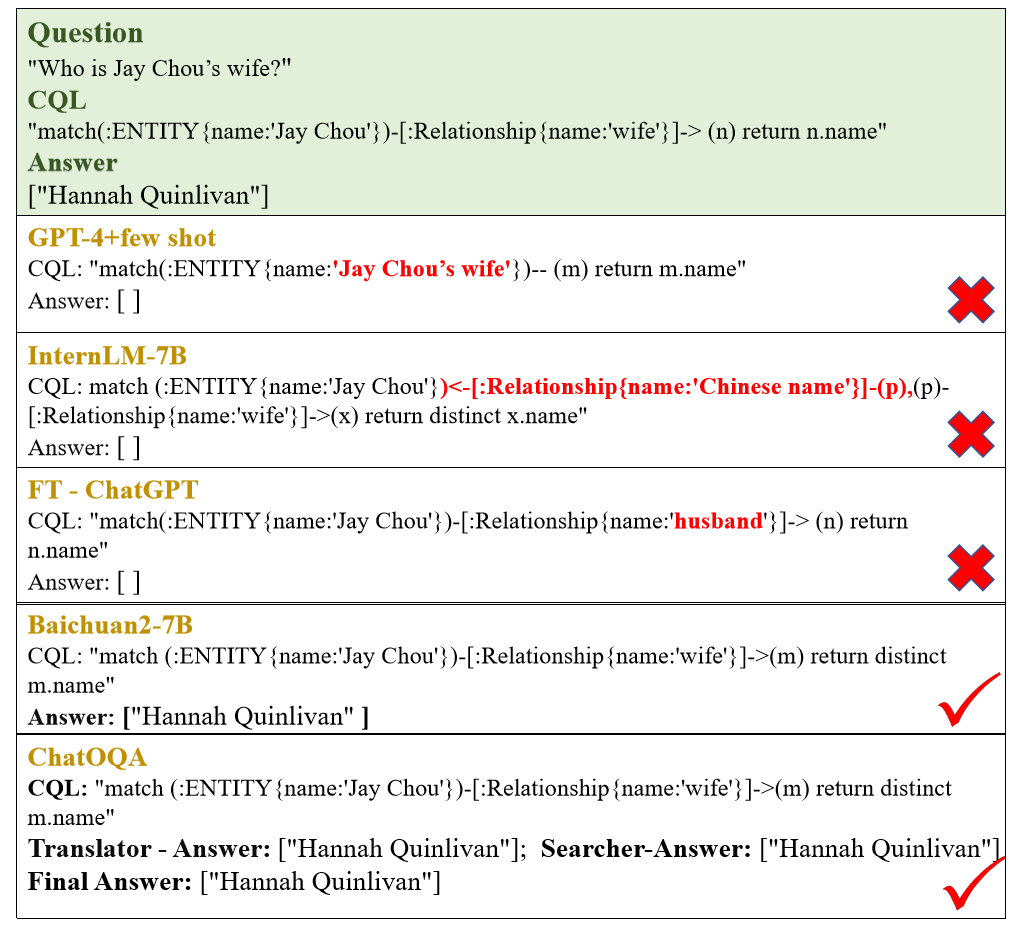}
  \caption{Qualitative examples from test results. The red areas indicate the parts that lead to errors. The original text is in Chinese, and we have translated the question, answer, and entities and relations into English.}
  \label{fig:case}
\end{figure}

Qualitatively, Figure \ref{fig:case} shows 5 examples generated by the baselines, and UniOQA. 
We carefully selected one of the most representative questions from the test results. 
It can be seen that models with relatively poor performance struggle to generate correct CQL, often containing grammatical errors or issues with entities and relations. 
On the other hand, models with better performance can generate correct CQL. 
Our framework can produce results both through execution and retrieval. 
The combination of these two methods ensures accuracy.

 In summary, through extensive experiments on the SpCQL, we draw the following conclusions: 
 (1) UniOQA exhibits a significant improvement over existing methods on SpCQL, achieving the new state-of-the-art performance. 
 (2) On the SpCQL dataset, FT-based models outperform non-parameter updated models (ICL-based and COT-based) by a significant margin. 
 (3) Among all FT-based models, Baichuan2-7B performs the best on SpCQL.

\subsection{Ablation Study}
We have conducted a macroscopic analysis of whether UniOQA improves the accuracy of question answering. 
In this section, we will answer the following questions: 
(1) How can we enhance the representation capacity of language models?
(2) Why are execution-based methods difficult to improve accuracy?
(3) What is the best way to combine the \textit{\textsf{Translator}} workflow and the \textit{\textsf{Searcher}} workflow?
(4) Should we adopt a unified framework for question answering?
\subsubsection{
{\bfseries How can we enhance the representation capacity of language models? }}
Strong representation capability means the model can fully understand the semantics of the problem.
To alleviate the overfitting of the fine-tuned LLM and thus enhance the model's semantic understanding ability, we utilize a method called Neftune (Neft)\cite{neft} in Section \ref{sec:ftllm}, which introduces some uniformly distributed noise into the embedding vectors. 
From Table \ref{tab:approach}, we observe a slight improvement. 
This indicates that {\bfseries Neft deepens the model's understanding of CQL.}  
However, our improvement is limited since we only add simple noise to the embedding values.
\begin{table}[htbp]
  \caption{Performance comparison of UniOQA without various modules(\%)}
  \label{tab:approach}
  \begin{tabular}{lccccc}
    \toprule
    & \multicolumn{4}{c}{\textbf{SpCQL}}  \\ \cline{2-5} 
    \multirow{-2}{*}{\textbf{Module}}& $\mathbf{ACC_{EX}}$ & $\mathbf{P}$ & $\mathbf{R}$ & $\mathbf{F1}$ \\
    \midrule
  w/o ERR & 53.7 & 64.5 & 65.8 & 63.7 \\
  w/o Neft & 54.8 & 66.8 & 69.0 & 66.1 \\
  w/o Neft+ERR & 53.2 &64.2&65.6&63.4\\
    \midrule
   UniOQA(ours) & $\textbf{54.9}$ & $\textbf{67.1}$ & $\textbf{69.2}$ & $\textbf{66.3}$ \\
    \bottomrule
  \end{tabular}
\end{table}

Strong representation capability also means the model can accurately capture entities and relations. The information source of the generated CQL is only the natural language questions if we miss alignment with knowledge graphs. 
Therefore, we introduce the Entity and Relation Replacement (ERR) algorithm (Section \ref{sec:algr}) to modify the entities and relations in CQL. 
We quantify the exact impact of ERR. 
As shown in Table \ref{tab:approach}, ERR significantly improves the output performance. 
The result indicates that {\bfseries ERR efficiently modifies the CQL without correct entities and relations.} However, CQL may have multiple relations that require replacement, and we currently only replace the first relation, leaving some room for improvement.

\subsubsection{
  \label{sec:why}
{\bfseries Why are execution-based methods difficult to improve accuracy?}}
 Recall in Section \ref{sec:introdction},  we have discussed why execution-based methods are difficult to further improve the overall question answering accuracy.  
 In this section, we verify the existence of this bottleneck by varying the training scale of the LLM in the \textit{\textsf{Translator}} workflow.

\begin{figure}[h]
  \centering
  \includegraphics[width=\linewidth]{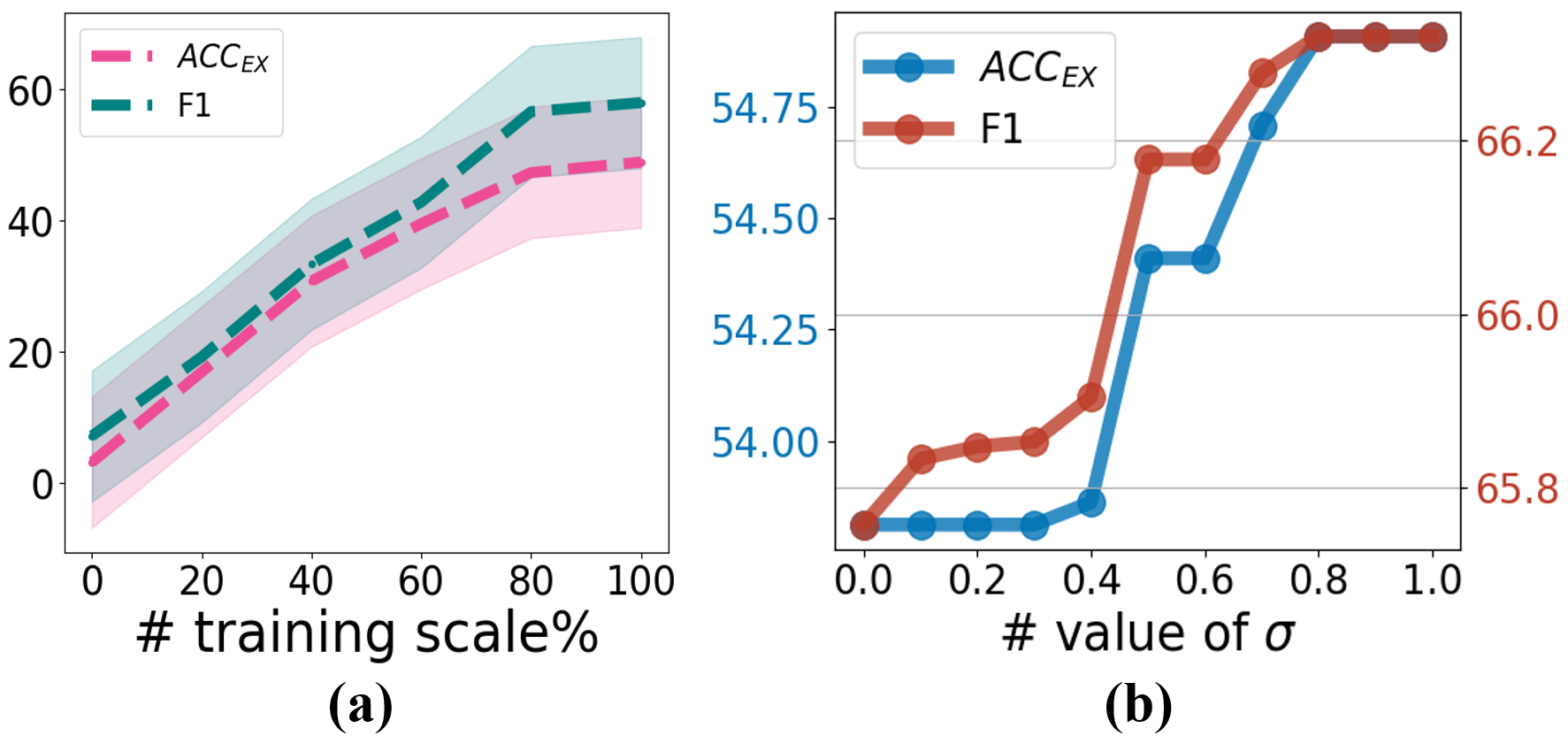} 
  \caption{(a) Performance of models trained on different scales (datasize \& step); (b) Performance of UniOQA as the decision operator $\sigma$ changes.}
  \label{fig:abla}
\end{figure}

We fine-tune Baichuan2-7B with the training scale of 20\% (1400 samples and 800 training steps), 40\% (2800 samples and 1600 training steps), 60\% (4200 samples and 2400 training steps), and 80\% (5600 samples and 3200 training steps). We utilize 100\% scale fine-tuning (7001 samples and 4000 training steps) as the baseline.
As shown in Figure \ref{fig:abla}(a), we observe that the effect improves with increased training volume, demonstrating the efficacy of fine-tuning. 
However, when it reaches a certain threshold, the model approaches convergence. At this point, {\bfseries attempting to directly optimize the model by expanding the training scale becomes prohibitively expensive.}  
This is precisely where the bottleneck lies.

\subsubsection{
{\bfseries What is the best way to combine the \textit{\textsf{Translator}} workflow and the \textit{\textsf{Searcher}} workflow? }}
Our two complementary workflows generate results in parallel, and are combined to generate the final results. 
In Section \ref{sec:comb}, we introduce $\sigma$, which is a decision factor that needs to be determined. 

Intuitively, as $\sigma$ approaches 0, the framework tends to favor the answers of \textit{\textsf{Translator}}. And the \textit{\textsf{Searcher}}'s results are only considered when the \textit{\textsf{Translator}} returns an empty result. As $\sigma$ approaches 1, the framework tends to select the better answers between the \textit{\textsf{Translator}} and the \textit{\textsf{Searcher}}.

From Figure \ref{fig:abla}(b), it can be observed that $\sigma = 1$ gives the best results, which means that {\bfseries in each of the two workflows, we select the superior answer as the final answer.}
Moreover, the performance is worse when $\sigma$=0, indirectly demonstrating that the dynamic decision algorithm outperforms previous ones\cite{decaf} as we equally consider the results of execution and retrieval.

\subsubsection{
{\bfseries Should we adopt a unified framework for question answering?}}
We have discussed the accuracy bottleneck of \textit{\textsf{Translator}} in Section \ref{sec:why}. 
 UniOQA combines the \textit{\textsf{Searcher}} workflow to further improve the accuracy of question answering.
As Table \ref{tab:flow} shown, {\bfseries the combination of the two workflows yields the best results, compared to running \textit{\textsf{Translator}} alone and \textit{\textsf{Searcher}} alone.} 
Intuitively, this is attributed to the complementary advantages of the two workflows, which mitigate each other's shortcomings.  

\begin{figure}[h]
  \centering
  \includegraphics[width=\linewidth]{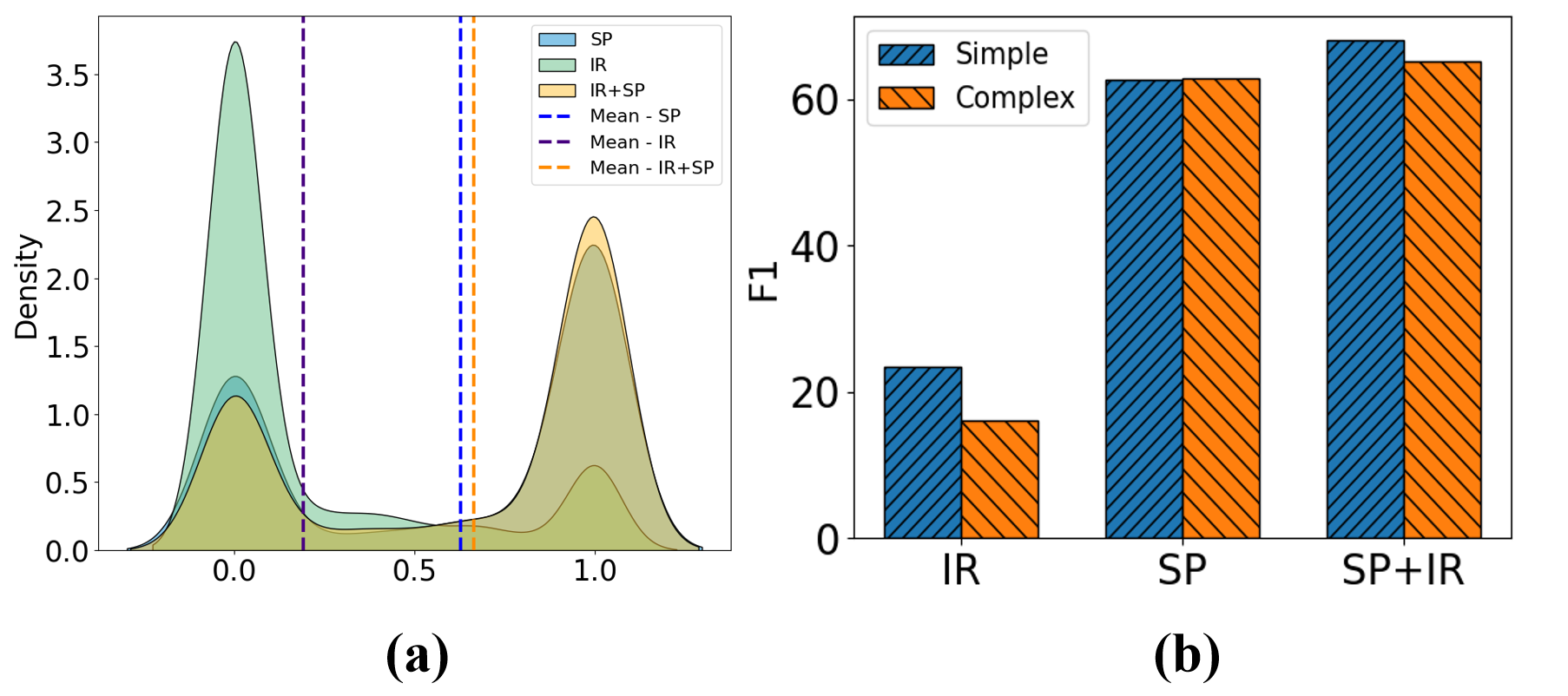}
  \caption{(a) Distribution of results across unique workflows; (b) Performance comparison of different workflows under varying difficulty levels. 
  In the two figures, SP refers to the \textit{\textsf{Translator}} workflow, and IR refers to the \textit{\textsf{Searcher}} workflow.}
  \Description{3 workflow in previous that generates wrong CQL.}
  \label{fig:density}
\end{figure}

To quantitatively explain this phenomenon, we categorize the test set questions into complex and simple ones based on whether they contain conditional clauses such as COUNT(*), WHERE, ORDER BY, or involve multiple entities and relations. 
There are 1142 complex questions and 865 simple questions. 
Subsequently, we evaluate the performance of the three workflows on these two types of questions. 
As shown in the Figure \ref{fig:density} (b), 
\textit{\textsf{Translator}} provides stable and accurate answers for both simple and complex questions. 
After combining with \textit{\textsf{Searcher}}, the accuracy of the question answering has been further improved. However, since the \textit{\textsf{Searcher}} workflow performs relatively poorly on complex questions, the improvement is limited.

\begin{table}[htbp]
  \caption{Performance comparison of various workflows(\%)}
  \label{tab:flow}
  \begin{tabular}{lccccc}
    \toprule
    & \multicolumn{4}{c}{\textbf{SpCQL}}  \\ \cline{2-5} 
    \multirow{-2}{*}{\textbf{Workflow(s)}}& $\mathbf{ACC_{EX}}$ & $\mathbf{P}$ & $\mathbf{R}$ & $\mathbf{F1}$ \\
    \midrule
    \textit{\textsf{Translator}} & 51.9 & 63.3 & 65.5 & 62.8 \\
    \textit{\textsf{Searcher}} & 11.7 & 20.3 & 21.6 & 19.3 \\
    \textit{\textsf{Translator}}+\textit{\textsf{Searcher}} &$\textbf{54.9}$ & $\textbf{67.1}$ & $\textbf{69.2}$ & $\textbf{66.3}$ \\
    \bottomrule
  \end{tabular}
\end{table}

We further analyze the distribution of each workflow results. 
As shown in Figure \ref{fig:density}(a), the F1 scores of \textit{\textsf{Searcher}} has a higher density distribution at smaller values, while \textit{\textsf{Translator}} has a higher density distribution at larger values. 
Moreover, the results of \textit{\textsf{Translator}}+\textit{\textsf{Searcher}} (i.e., UniOQA) are most densely distributed at larger values and have the highest mean, further verifying the superiority of the unified framework. 

\section{CONCLUSION}

In this paper, to address the challenges of inadequate model representation capacity and the difficulty in further improving the question answering accuracy, we propose UniOQA, a unified framework combined with two parallel workflows. 
It can be observed that by fine-tuning large language models to generate CQL and further replacing entities and relations, we can effectively comprehend the semantics of the questions and generate correct CQL.  
Additionally, we transfer the Retrieval-Augmented Generation process to the knowledge graph, enhancing the overall accuracy of question answering.  
Experimental results demonstrate that we significantly improve 
SpCQL Logical Accuracy and Execution Accuracy, achieving the new state-of-the-art results on SpCQL.  
Subsequent ablation studies further analyze the superiority of UniOQA
in representation capacity and quantify its breakthrough in performance. 
 Our work provides valuable theoretical and practical insights for knowledge graph question answering.

In future work, we will further explore question answering systems that effectively combine unstructured data with large language models. We will also focus on balancing the model's efficiency and resource consumption.

\bibliographystyle{ACM-Reference-Format}
\bibliography{sample-base}

\end{document}